\documentclass{article}

\usepackage{microtype}
\usepackage{graphicx}
\usepackage{subfigure}
\usepackage{booktabs} 

\usepackage{hyperref}

\usepackage[accepted]{icml2023}

\usepackage{amsmath}
\usepackage{amssymb}
\usepackage{mathtools}
\usepackage{amsthm}

\usepackage[capitalize,noabbrev]{cleveref}

\theoremstyle{plain}

\theoremstyle{definition}

\theoremstyle{remark}

\usepackage[textsize=tiny]{todonotes}

\icmltitlerunning{Submission and Formatting Instructions for ICML 2023}

\begin{document}

\twocolumn[
\icmltitle{Speech Wikimedia: A 77 Language Multilingual Speech Dataset}



\icmlsetsymbol{equal}{*}

\begin{icmlauthorlist}
\icmlauthor{Rafael Mosquera Gómez}{equal,Factored.ai}
\icmlauthor{Julian Eusse}{equal,Factored.ai}
\icmlauthor{Juan Ciro}{Factored.ai}
\icmlauthor{Daniel Galvez}{nvidia}
\icmlauthor{Ryan Hileman}{talon}
\icmlauthor{Kurt Bollacker}{longnow}
\icmlauthor{David Kanter}{mlcommons}
\end{icmlauthorlist}

\icmlaffiliation{Factored.ai}{Factored.ai}
\icmlaffiliation{nvidia}{NVIDIA}
\icmlaffiliation{talon}{Talon Voice}
\icmlaffiliation{longnow}{Long Now Foundation}
\icmlaffiliation{mlcommons}{MLCommons}


\icmlcorrespondingauthor{Daniel Galvez}{dgalvez@nvidia.com}

\icmlkeywords{Machine Learning, ICML}

\vskip 0.3in
]



\printAffiliationsAndNotice{\icmlEqualContribution} 
\begin{abstract}
The Speech Wikimedia Dataset is a publicly available compilation of audio with transcriptions extracted from Wikimedia Commons. It includes 1780 hours (195 GB) of CC-BY-SA licensed transcribed speech from a diverse set of scenarios and speakers, in 77 different languages. Each audio file has one or more transcriptions in different languages, making this dataset suitable for training speech recognition, speech translation, and machine translation models.

\end{abstract}

\section{Introduction}
\label{intro}

Speech research has witnessed remarkable advancements in recent years, largely driven by the availability of vast amounts of data. Tasks such as automatic speech recognition (ASR), speaker recognition, and speech translation have reached robust results even in the presence of background noise, jargon, and different accents. Nevertheless, one of the fundamental challenges in speech research is the scarcity of multilingual datasets. 

In this paper, we introduce the Speech Wikimedia Dataset, a supervised dataset consisting of 1780 hours of audio files, each with one one or more transcripts. The last point bears repeating: much (approximately 25\%) of the dataset has multiple associated transcripts, each having its own language, which is rare in datasets sourced from the internet.
The dataset is specifically curated from Wikimedia Commons to address some of the key challenges in the speech recognition, machine translation, and speech translation spaces, particularly the need for diverse multilingual data and appropriate licensing for academic and commercial usage.

The paper is organized as follows. In section 2, we describe the dataset itself and training tasks that it can support; in section 3, we describe related work; in section 4, we describe some limitations of the dataset today; in section 5, we conclude with some future work that this dataset can enable.

\section{Dataset Description}
\label{description}

To construct the Speech Wikimedia dataset, we downloaded raw video and audio from Wikimedia Commons (``https://commons.wikimedia.org''), which allows only content that is "free"  \cite{wikimedia_license}. For our purposes, this means data under CC-BY or CC-BY-SA license, or otherwise public domain. After downloading, the data was converted to 16kHz monochannel FLAC using ffmpeg. The data is uploaded to Huggingface at \url{https://huggingface.co/datasets/MLCommons/speech-wikimedia}

We give statistics for three possible tasks that this dataset can be used for: speech recognition, speech translation, and machine translation.

\subsection{Licensing Information}

Given that all data is public domain, CC-BY-licensed, or CC-BY-SA licensed, we are licensing the dataset as CC-BY-SA. Following the requirements imposed by the CC-BY and CC-BY-SA licenses of our sources, accreditation is provided in the linked \href{https://huggingface.co/datasets/MLCommons/speech-wikimedia/blob/main/credits.json}{credits.json} file.

\subsection{Audio with Subtitles in the Same Language for Speech Recognition Task}

In order to determine the amount of data available for the ASR task we used only those audios and transcriptions where the language coincided. Since Wikimedia Commons's transcripts' filenames contain the language of the contained text, we simply extracted the languages from these filenames. For example, the file ``Elephants\_Dream.ogv.en.srt'' is in English, as indicated by the ``.en.'' substring.

Given that we didn't initially have the audio language for each file, Whisper's \cite{radford2022robust} language detection pipeline was used. A total of 77 different languages were detected, with English, Dutch, German, Russian, and Spanish being the most common. 69.07\% of the 1780 hour dataset comprises audio-transcription pairs in the same language. We present the number of transcribed hours of each language in Table 1 in the appendix.

\subsection{Audio with Subtitles in Different Languages for Speech Translation Task}

For speech translation, we focused on audio-transcript pairs that had different languages, which corresponds to 31\% of the rest of the 1780 hours. After filtering audios and transcriptions with unknown languages, we were left with a total of 628.8 hours of audio with transcripts in different languages. We present the hours of audio for the 20 most common language pairs in Table 2 in the appendix.

\subsection{Transcript Language Pairs (Bitexts) for Machine Translation Task}

While the speech translation task relies on audio from one language with a transcription from another language, machine translation focuses on pure text translation. We find paired texts by enumerating all pairs of transcripts associated with a single audio. This is interesting in particular because approximately 10.93\% of the  audio files have transcripts in at least three languages. 

In Table 3 in the appendix, we present total hours, number of words and occurrences of different pairs of languages for the 20 most common language pairs in the dataset.

\subsection{Topic Distribution and Audio Content}

We were also interested in determining which topics were covered across the dataset. In order to analyze this, we ran a zero shot classifier \cite{ktrain} with labels ranging different topics, and recorded the hours of audio for each of the topics. Results are depicted in Table 4 in the Appendix section. Popular topics were current events, history, and general non-fiction references.

Based on listening to several files, we also discovered that several audios are public speeches, music, and clearly pronounced single words, probably for pronunciation dictionaries like wiktionary.

\section{Related Work}
\label{related}

In this section we provide an overview of previous, similar datasets.

\textbf{Mozilla Common Voice} \cite{ardila2020common} is a CC0-licensed 17,690-hour public domain corpus of single speaker read speech in 108 languages created by volunteers. In contrast, Speech Wikimedia has much more diverse audio sources.


\textbf{Multilingual Librispeech} \cite{Pratap_2020} is a CC-BY dataset of 50,500  hours of transcribed read speech in eight languages; 6,000 of its hours are non-English. Meanwhile, our dataset contains 77 languages, and the majority of the data is also in English.

\textbf{VoxPopuli} \cite{DBLP:journals/corr/abs-2101-00390} is CC0-licensed dataset containing an unsupervised set of 400,000 hours in 23 languages, and 1,500 hours of transcribed audio in 15 languages. Like our dataset, it also contains a subset suitable for a speech translation task.

\textbf{Multilingual Spoken Words Corpus} \cite{mazumder2021multilingual} is a CC-BY licensed 6,000-hour dataset, containing more than 340,000 keywords in 50 different languages. It is for training keyword spotting models, not speech recognition models.

\textbf{opensubtitles} \cite{lison-tiedemann-2016-opensubtitles2016} is a machine translation dataset containing 1,782 language pairings extracted from movie subtitles in 62 languages. Given the data source, it is not licensed for commercial usage. In contrast, the Speech Wikimedia Dataset has 929 language pairings from 77 languages.

\section{Limitations}

The raw data is available publicly online on Hugging Face as mentioned before; however, this data is not yet processed via forced alignment of audio to transcript and bitext word alignment for transcript to transcript, and thus not able to be used immediately in training models.

We removed all video data when converting to FLAC. In future work, this data could be helpful for a multimodal task.

While collecting this dataset, we realized that there is also a collection of audio data in Wikimedia Commons without any transcripts. We have not explored this subset and have not made it available at this time, however.

Given the small size of the dataset, we are not providing a training-test split.


\section{Conclusions}
We introduce the Speech Wikimedia Dataset, a collection of audio files with transcriptions in multiple languages extracted form Wikimedia Commons. The dataset encompasses over 1,780 hours of transcribed speech in multiple languages. The CC-BY-SA license enables commercial usage. This is the first non-read multilingual speech dataset allowing for commercial usage that we are aware of other than VoxPopuli.






\bibliography{speech_wikimedia_bib}
\bibliographystyle{icml2023}

\newpage
\appendix
\onecolumn
\section{Figures}

\begin{table}[t]
\caption{Automatic Speech Recognition Task }
\label{sample-table}
\vskip 0.15in
\begin{center}
\begin{small}
\begin{sc}
\begin{tabular}{lcccr}
\toprule
{} &  Hours of audio \\
Language &                 \\
\midrule
English (en)           &     1488.765773 \\
Dutch (nl)           &       22.167223 \\
German (de)           &       12.658670 \\
French (fr)           &        7.163889 \\
Russian (ru)           &        6.985941 \\
Spanish (es)           &        6.184720 \\
Latin (la)           &        3.066669 \\
Polish (pl)           &        3.045028 \\
Japanese (ja)           &        2.216300 \\
Bengali (bn)           &        2.126192 \\
Swedish (sv)           &        1.468487 \\
Chinese (zh)           &        1.456599 \\
Italian (it)           &        1.419221 \\
Portuguese (pt)           &        1.344584 \\
Welsh (cy)           &        1.141955 \\
Basque (eu)           &        1.008435 \\
Hindi (hi)           &        0.886795 \\
Arabic (ar)           &        0.572991 \\
Ukrainian (uk)           &        0.441770 \\
Slovenian (sl)           &        0.377644 \\
Korean (ko)           &        0.367545 \\
Hebrew (he)           &        0.238240 \\
Indonesian (id)           &        0.207363 \\
Thai (th)           &         0.177196 \\
Catalan (ca)           &        0.161531 \\
Greek (el)           &        0.160628 \\
Danish (da)           &        0.150981 \\
Persian (fa)           &        0.132622 \\
Vietnamese (vi)           &        0.131922 \\
Marathi (mr)           &        0.124219 \\
Punjabi (pa)           &        0.090774 \\
Malayalam (ml)           &        0.078354 \\
Telugu (te)           &        0.065369 \\
Kannada (kn)           &        0.033602 \\
Hungarian (hu)           &        0.030055 \\
Estonian (et)           &        0.029325 \\
Turkish (tr)           &        0.024743 \\
Finnish (fi)           &        0.022719 \\
Czech (cs)           &        0.021120 \\
Telugu (tl)           &        0.016138 \\
Romanian (ro)           &        0.015280 \\
Slovak (sk)           &        0.000766 \\
Tamil (ta)           &        0.000364 \\
\bottomrule
\end{tabular}
\label{tab:total-time}
\end{sc}
\end{small}
\end{center}
\vskip -0.1in
\end{table}
\begin{table}[t]
\caption{Speech Translation Task }
\label{sample-table}
\vskip 0.15in
\begin{center}
\begin{small}
\begin{sc}
\begin{tabular}{lcccr}
\toprule
Audio Language & Transcript Language & Duration(hours) \\
\midrule
English & Spanish & 67.115705 \\
English & Arabic & 43.398845 \\
English & French & 38.163062 \\
English & Portuguese & 30.952778 \\
English & Dutch & 24.165356 \\
English & German & 23.678866 \\
English & Italian & 23.442334 \\
English & Russian & 15.557022 \\
Dutch & English & 14.409074 \\
English & Polish & 12.865772 \\
Latin & English & 11.722308 \\
English & Chinese & 11.182589 \\
Hindi & English & 10.256298 \\
English & Turkish & 9.471801 \\
English & Japanese & 8.782186 \\
Welsh & English & 8.761795 \\
English & Vietnamese & 6.731008 \\
Russian & English & 6.037366 \\
Dutch & Russian & 5.438943 \\
\bottomrule
\end{tabular}
\label{tab:total-time}
\end{sc}
\end{small}
\end{center}
\vskip -0.1in
\end{table}
\begin{table}[t]
\caption{Transcript Language Pairs Statistics }
\label{sample-table}
\vskip 0.15in
\begin{center}
\begin{small}
\begin{sc}
\begin{tabular}{lcccr}
\toprule
Language Pair & Total Hours & Source Language Token Count & Target Language Token Count & Bitexts \\
\midrule
English-Spanish & 135.989042 & 481391.0 & 486965.0 & 629 \\
English-French & 85.782796 & 262040.0 & 255998.0 & 343 \\
English-Portuguese & 57.887887 & 200853.0 & 194911.0 & 197 \\
English-Russian & 55.501208 & 149706.0 & 119638.0 & 348 \\
German-English & 55.449766 & 149499.0 & 168156.0 & 394 \\
Spanish-Portuguese & 54.185978 & 200486.0 & 191356.0 & 166 \\
Spanish-French & 51.878961 & 178583.0 & 182886.0 & 213 \\
English-Dutch & 49.582888 & 182302.0 & 166567.0 & 164 \\
English-Italian & 47.008579 & 131800.0 & 125312.0 & 200 \\
French-Portuguese & 38.802198 & 147000.0 & 138013.0 & 146 \\
Arabic-English & 38.239120 & 106115.0 & 136589.0 & 182 \\
German-Spanish & 36.046692 & 110171.0 & 127857.0 & 211 \\
Arabic-Spanish & 34.548516 & 109102.0 & 136234.0 & 139 \\
Arabic-French & 34.121227 & 110543.0 & 138088.0 & 134 \\
German-French & 33.843628 & 94528.0 & 111353.0 & 204 \\
French-Italian & 33.791085 & 117284.0 & 113286.0 & 150 \\
Spanish-Italian & 33.368969 & 117633.0 & 109450.0 & 162 \\
Arabic-Portuguese & 29.675284 & 96408.0 & 113835.0 & 98 \\
German-Italian & 28.917809 & 85169.0 & 96215.0 & 154 \\
French-Russian & 27.784403 & 80372.0 & 63862.0 & 155 \\
\bottomrule
\end{tabular}
\label{tab:total-time}
\end{sc}
\end{small}
\end{center}
\vskip -0.1in
\end{table}
\begin{table}[t]
\caption{Distribution of topics and their durations}
\label{sample-table}
\vskip 0.15in
\begin{center}
\begin{small}
\begin{sc}
\begin{tabular}{lcccr}
\toprule
Topic &  Duration(hours)\\
\midrule
current events             &      641.809422 \\
other             &      406.496021 \\
history           &      154.644719 \\
health            &      151.203017 \\
general reference &      114.263664 \\
society           &       95.286515 \\
political         &       46.760335 \\
technology        &       46.079005 \\
number            &       45.406569 \\
business          &       40.940404 \\
science           &       34.944243 \\
culture           &       27.520957 \\
languages         &       26.272240 \\
city              &       24.718620 \\
location          &       15.258170 \\
software          &        8.970613 \\
geography         &        8.372475 \\
animal            &        7.417614 \\
religion          &        7.382679 \\
philosophy        &        6.863642 \\
art               &        5.678405 \\
entertainment     &        5.076250 \\
mathematics        &        2.186313 \\
crypto            &        1.731447 \\
gaming              &        1.252531 \\
engineering          &        0.154807 \\
\bottomrule
\end{tabular}
\label{tab:total-time}
\end{sc}
\end{small}
\end{center}
\vskip -0.1in
\end{table}


\end{document}